\documentclass{article}

\usepackage{microtype}
\usepackage{graphicx}
\usepackage{subcaption}
\usepackage{booktabs} 
\usepackage{soul}
\usepackage{pifont}
\usepackage{hyperref}

\usepackage[preprint]{icml2026}

\usepackage{amsmath}
\usepackage{amssymb}
\usepackage{mathtools}
\usepackage{amsthm}
\usepackage{multicol}
\usepackage{multirow}

\usepackage[capitalize,noabbrev]{cleveref}

\theoremstyle{plain}

\theoremstyle{definition}

\theoremstyle{remark}

\usepackage[textsize=tiny]{todonotes}

\newcommand{\n}{\texttt{FedMOA}}

\newcommand{\eg}{\textit{e.g., }}
\newcommand{\ie}{\textit{i.e., }}

\icmltitlerunning{FedMOA: Federated GRPO for Personalized Reasoning LLMs under Heterogeneous Rewards}

\begin{document}


\twocolumn[
  \icmltitle{FedMOA: Federated GRPO for Personalized Reasoning LLMs under Heterogeneous Rewards}



  \icmlsetsymbol{equal}{*}

  \begin{icmlauthorlist}
    \icmlauthor{Ziyao Wang}{equal,yyy}
    \icmlauthor{Daeun Jung}{equal,yyy}
    \icmlauthor{Yexiao He}{yyy}
    \icmlauthor{Guoheng Sun}{yyy}
    \icmlauthor{Zheyu Shen}{yyy}
    \icmlauthor{Myungjin Lee}{comp}
    \icmlauthor{Ang Li}{yyy}
  \end{icmlauthorlist}

  \icmlaffiliation{yyy}{University of Maryland, College Park}
  \icmlaffiliation{comp}{Cisco Research}

  \icmlcorrespondingauthor{Ziyao Wang}{ziyaow@umd.edu}
  \icmlcorrespondingauthor{Ang Li}{angliece@umd.edu}

  \icmlkeywords{Machine Learning, ICML}

  \vskip 0.3in
]



\printAffiliationsAndNotice{\icmlEqualContribution}

\begin{abstract}
Group Relative Policy Optimization (GRPO) has recently emerged as an effective approach for improving the reasoning capabilities of large language models through online multi-objective reinforcement learning. 
While personalization on private data is increasingly vital, traditional Reinforcement Learning (RL) alignment is often memory-prohibitive for on-device federated learning due to the overhead of maintaining a separate critic network. GRPO's critic-free architecture enables feasible on-device training, yet transitioning to a federated setting introduces systemic challenges: heterogeneous reward definitions, imbalanced multi-objective optimization, and high training costs.
We propose \textbf{{\n}}, a federated GRPO framework for multi-objective alignment under heterogeneous rewards. 
{\n} stabilizes local training through an online adaptive weighting mechanism via hypergradient descent, which prioritizes primary reasoning as auxiliary objectives saturate. On the server side, it utilizes a task- and accuracy-aware aggregation strategy to prioritize high-quality updates.
Experiments on mathematical reasoning and code generation benchmarks demonstrate that {\n} consistently outperforms federated averaging, achieving accuracy gains of up to 2.2\% while improving global performance, personalization, and multi-objective balance.
\end{abstract}

\section{Introduction}
Online reinforcement learning (RL)~\cite{kaelbling1996reinforcement} is increasingly used to improve the reasoning capabilities of large language models (LLMs) beyond what supervised pre-training and fine-tuning can provide~\cite{yu2025dapo}. In particular, Group Relative Policy Optimization (GRPO)~\cite{guo2025deepseek} has shown that carefully designed online RL pipelines can substantially boost performance on mathematically intensive problems, code generation, and other structured reasoning tasks, and is now a standard component in many commercial LLM APIs. However, these GRPO pipelines are almost exclusively studied and deployed in a centralized setting, where a single trainer performs post-training optimization on a globally collected dataset with a unified reward pipeline to obtain one global reasoning-oriented model. As demand for personalized and agentic LLMs grows~\cite{sun2025fedagent} , this centralized paradigm is increasingly limited by the coverage and diversity of centrally available data. Effective personalization requires direct use of user interaction data for reasoning alignment, yet such data often containing sensitive corporate codebases or private interaction logs cannot be shared with a central trainer due to privacy or regulatory constraints~\cite{yao2024survey,yan2024protecting,yan2025protecting,wang2025prada}.

\begin{figure}[t]
    \centering  \includegraphics[width=2.5in]{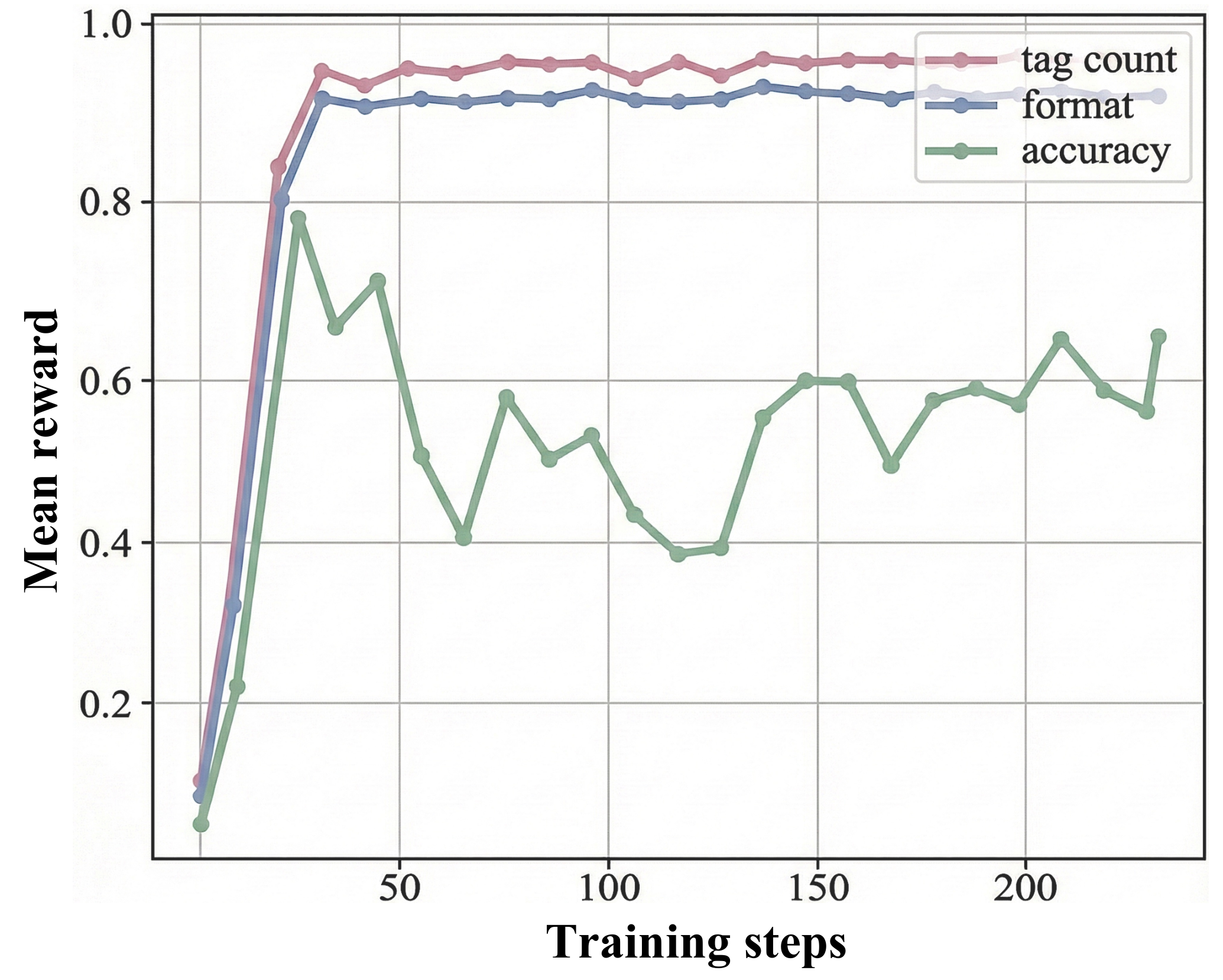}
    \caption{Training curves of multi-objective rewards in federated GRPO. Format-related rewards converge rapidly, while accuracy-related rewards remain low and exhibit substantial fluctuations.}
    \label{fig:cost}
\end{figure}

Federated learning (FL)~\cite{mcmahan2017fedavg,li2018fedprox} offers a natural way to mitigate these limitations by enabling joint training without sharing raw data. 
While FL is effective for supervised fine-tuning of LLMs~\cite{kuang2024federatedscope,ye2024openfedllm,wang2024flora}, \ul{transitioning from centralized to federated GRPO is not a trivial drop-in change, it is a fundamental reconfiguration of the multi-objective optimization landscape.} This introduces three key challenges.
\ding{182}\emph{Heterogeneous rewards}: clients may focus on different domains and tasks (\eg some clients focus on mathematical problems, others on code, others on medical content)~\cite{wang2020tackling,luo2021no}, leading to misaligned local reward signals. 
\ding{183}\emph{Multi-objective alignment}: prior work has observed that GRPO can exhibit imbalanced progress across objectives (\eg accuracy, format, tags)~\cite{lu2025learning}, and in FL, this imbalance is exacerbated by limited per-client data and the difficulty of tuning hyperparameters in a distributed setting, as shown in Fig~\ref{fig:cost}. 
\ding{184}\emph{Computational cost}: GRPO rollout remains expensive, requiring a scheme that achieves convergence in few communication rounds to remain resource-feasible~\cite{konevcny2016federated,shahid2021communication}.

To address these challenges, we propose \textbf{{\n}}, a \underline{Fed}erated GRPO algorithm with \underline{M}ulti-\underline{O}bjective \underline{A}lignment. 
Each client specifies a local multi-objective reward function reflecting its GRPO training goals. During local training, the client dynamically adjusts the associated reward weights via a hypergradient-based optimization scheme~\cite{baydin2017online}: objectives that are already well optimized are down-weighted, while under-optimized objectives receive increased emphasis~\cite{baydin2017online}.
At the end of each communication round, clients transmit their updated model parameters together with the adapted reward weights to the server. 
The server groups clients by task and performs progress-aware aggregation within each task group, where clients that exhibit stronger progress on the shared accuracy-related objective, as reflected by their adapted reward weights, contribute more to the aggregated update. 
After task-wise aggregation, the server applies a standard aggregation step across tasks and redistributes the aggregated model parameters to the corresponding clients for the next communication round.
This design allows each client to autonomously learn how to balance local objectives while using reward-derived signals to guide global aggregation without requiring auxiliary validation data.


We summarize our contributions as follows:
\begin{itemize}
    \item We propose {\n}, the first federated GRPO algorithm for multi-objective alignment that enables privacy-preserving reasoning personalization through online RL under heterogeneous rewards.

    \item We identify three key challenges in federated GRPO, namely heterogeneous rewards, multi-objective alignment, communication and computational cost, and addresses them through adaptive objective weighting and progress-aware aggregation.

    \item Experiments on mathematical and coding benchmarks show that {\n} consistently improves global reasoning performance over federated baselines, achieving gains of up to 2.0\% on GSM8K, 2.2\% on MATH, and 1.8\% on HumanEval, while enabling more balanced multi-objective reward optimization under heterogeneous data–reward settings.
    
\end{itemize}

\section{Preliminaries}

\textbf{Multi-Objective Alignment with GRPO.} Multi-objective training considers policies that optimize multiple reward components simultaneously, rather than a single scalar reward~\cite{guo2024controllable,zhou2024beyond}. In our setting, as shown in Figure~\ref{fig:cost}, GRPO jointly optimizes an accuracy-related reward and auxiliary format-related rewards (\eg enforcing step-by-step reasoning or tag outputs), with the accuracy-related component serving as the primary objective, encouraging LLMs to produce both correct answers and well-structured outputs~\cite{guo2025deepseek}.

Formally, we assume $K$ objectives, each associated with a reward component $r_k$ and a nonnegative weight $w_k$, where $k \in \{1,\dots,K\}$,
\begin{equation}
    w_k \ge 0, \qquad \sum_{k=1}^K w_k = 1.
\end{equation}
The scalarized reward is
\begin{equation}
    r_w = \sum_{k=1}^K w_k r_k.
\end{equation}
Let $\pi_\theta$ denote the policy parameterized by $\theta$, and let $\tau$ denote a prompt-conditioned response group sampled from $\pi_\theta$. The multi-objective GRPO can be written as
\begin{equation}
    J(\theta; w) = \mathbb{E}_{\tau \sim \pi_\theta} \big[ R_w(\tau) \big],
    \quad
    R_w(\tau) = \sum_{t} r_w(s_t, a_t),
\end{equation}
and GRPO optimizes a Proximal Policy Optimization (PPO)-style~\cite{schulman2017proximal} policy-gradient surrogate using a group-relative advantage computed from the scalarized reward $r_w$.

\begin{equation}
    \max_{\theta} \; \mathbb{E}_{(s,a) \sim \pi_{\theta_{\mathrm{old}}}} 
    \big[ A_w(s,a)\, \log \pi_\theta(a \mid s) \big],
\end{equation}
where the group-relative advantage $A_w$ is computed by comparing rewards within each response group under the same prompt (\eg via within-group normalization) using the scalarized reward $r_w$.

\begin{figure}[t]
    \centering  
    \includegraphics[width=\columnwidth]{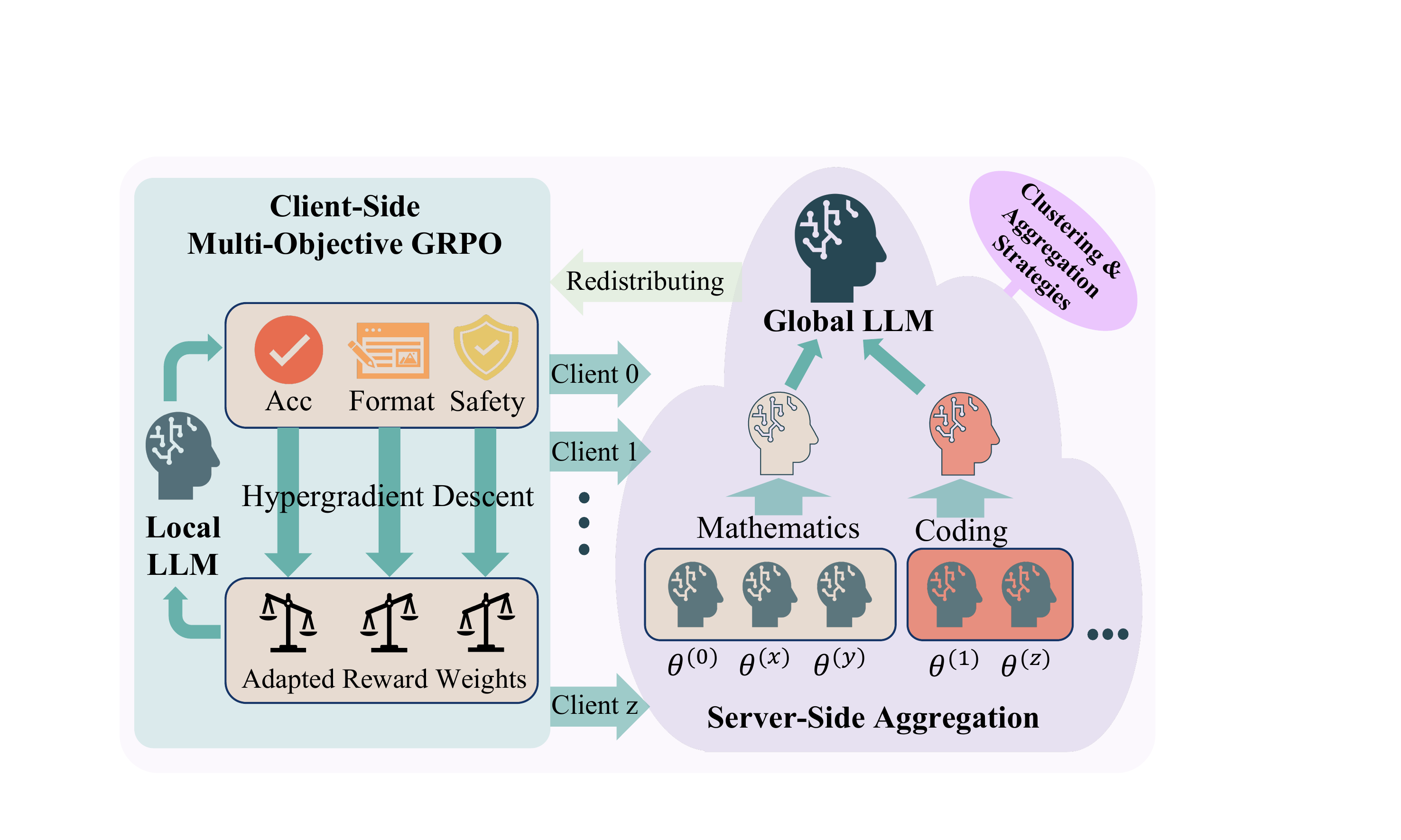}
    \caption{Overview of {\n}.
Each client performs multi-objective GRPO with a local hypergradient-based procedure to adapt objective weights, prioritizing accuracy while balancing auxiliary objectives.
The adapted weights are sent to the server together with model updates and used for task-aware grouping and accuracy-aware aggregation.
}
    \label{fig:overview}
\end{figure}

\textbf{Federated GRPO under Heterogeneous Rewards.}
In the FL setting, GRPO faces an additional layer of heterogeneity across clients. We consider $M$ clients indexed by $m \in \{1,\dots,M\}$. Each client $m$ is associated with its own set of $K_m$ objectives, with reward components $\{r_k^{(m)}\}_{k=0}^{K_m-1}$ and weights $\{w_k^{(m)}\}_{k=0}^{K_m-1}$, where in general $K_m$ and the semantics of the objectives may differ across clients. For each client, we assume that $r_0^{(m)}$ is directly tied to task accuracy (\ie correctness on domain-specific questions), and we treat $w_0^{(m)}$ as the weight of this accuracy-related objective. The remaining objectives may be accuracy-adjacent metrics (\eg precision or F1, when available) or format-related signals (\eg reasoning structure or style constraints), and can vary substantially across clients. This leads to inherently heterogeneous multi-objective reward functions, making it unclear how to directly aggregate client updates that are optimized under different scalarizations and even different sets of objectives. In practice, however, clients can often be grouped into coarse task categories, and within each category clients share a common accuracy-related objective $r_0^{(m)}$ that can be compared across clients, enabling aggregation schemes that use this shared notion of accuracy as a reference while keeping other objectives client-specific.

Within this federated GRPO setup, the three challenges from the introduction become concrete. First, the client-specific objective sets $\{r_k^{(m)}\}$ and weights $\{w_k^{(m)}\}$ make rewards inherently heterogeneous across clients, so naively averaging updates mixes gradients optimized under different scalarizations and even different sets of objectives. Second, even on a single client, multi-objective GRPO can exhibit imbalanced progress across objectives: as illustrated in Figure~\ref{fig:cost}, format-related rewards such as tag count and output structure quickly saturate near their maximum, while the accuracy reward plateaus at a much lower level and continues to fluctuate, suggesting that auxiliary objectives may dominate optimization early on and hinder improvements on the accuracy-related component if weights are not adapted. Third, GRPO itself is substantially more computationally expensive than supervised fine-tuning, and running it on many clients under tight resource constraints places a premium on algorithms that achieve good performance with few local steps and few communication rounds.

\section{Proposed Method: {\n}}
\subsection{Overview}
{\n} combines client-side adaptive objective weighting with task-aware server aggregation under heterogeneous rewards. Figure~\ref{fig:overview} shows the overview of {\n} workflow. On each client, we update weights during training via a hypergradient descent procedure: objectives whose gradients indicate continued progress are up-weighted, while those that appear close to convergence are down-weighted. In this way, the client iteratively adjusts its objective weights during local GRPO updates.
At the end of a local training round, each client sends its updated model parameters and adapted objective weights to the server. The server groups clients by task type (\eg math, coding, personal assistant) and, within each group, uses the adapted accuracy weights to prioritize clients that have made more progress on accuracy, assigning them larger aggregation weights. It then performs group-wise aggregation of model parameters and reward weights, followed by a FedAvg-style aggregation across groups based on data sizes, and finally redistributes the resulting global model and group-level reward weights to clients for the next round.

\subsection{Adaptive Objective Weighting via Hypergradient Descent}

We adapt hypergradient descent to update the objective weights on each client.
Classically, consider minimizing an empirical risk $F(w)$ with parameters $w$ and
a scalar learning rate $\eta^{(t)}$ at iteration $t$:
\begin{equation}
    w^{(t+1)} = w^{(t)} - \eta^{(t)} \nabla F\big(w^{(t)}\big).
\end{equation}
Hypergradient descent treats $\eta^{(t)}$ as a parameter and updates it using
the derivative of the next-step loss $F(w^{(t+1)})$ with respect to
$\eta^{(t)}$. Under a standard approximation, this derivative can be written as
an inner product between consecutive gradients,
\begin{equation}
    \frac{\partial F(w^{(t+1)})}{\partial \eta^{(t)}}
    \approx - \nabla F\big(w^{(t+1)}\big)^\top \nabla F\big(w^{(t)}\big),
\end{equation}
so that a gradient step on $\eta$ has the form
\begin{equation}
    \eta^{(t)} = \eta^{(t-1)} + \rho\,
    \nabla F\big(w^{(t)}\big)^\top \nabla F\big(w^{(t-1)}\big),
\end{equation}
where $\rho>0$ is a small step size. Intuitively, when consecutive gradients are
aligned (positive inner product), the learning rate is increased; when they
point in different directions (negative inner product), the learning rate is
decreased.

In our setting, each client $m$ treats its objective weights
$\{w_k^{(m)}\}_{k=0}^{K_m-1}$ as hyperparameters and updates them with a
similar rule. Let $L_k^{(m)}$ denote the loss associated with the $k$-th reward
component $r_k^{(m)}$. During GRPO training at iteration $t$, we compute, for
each objective $k$, a gradient with respect to an intermediate representation
$h^{(t)}$ (\eg a chosen hidden layer) rather than all model parameters,
\begin{equation}
    g_k^{(t)} = \nabla_{h} L_k^{(m)}\big(h^{(t)}\big).
\end{equation}
We then define a hypergradient signal as the inner product of consecutive
per-objective gradients,
\begin{equation}
    \Delta_k^{(t)} = g_k^{(t)}{}^\top g_k^{(t-1)},
\end{equation}
and update the corresponding weight by
\begin{equation}
    w_k^{(m,t)} = w_k^{(m,t-1)} + \lambda\, \Delta_k^{(t)},
\end{equation}
where $\lambda > 0$ is a hypergradient step size. After the update, the weight
vector is projected back onto the probability simplex to ensure nonnegativity
and $\sum_k w_k^{(m,t)} = 1$. A positive inner product $\Delta_k^{(t)}$ (stable,
aligned gradients) increases $w_k^{(m,t)}$, indicating that the objective is
still far from convergence and can be emphasized; a negative $\Delta_k^{(t)}$
reduces $w_k^{(m,t)}$, down-weighting objectives whose gradients oscillate or
are close to convergence. By computing $g_k^{(t)}$ only on a single
intermediate layer and for one reward component at a time, this procedure keeps
the additional cost of adaptive weighting modest relative to standard GRPO
training.

\subsection{Local-Reward-Guided Global Aggregation}

In the global phase, {\n} uses the locally adapted reward weights to guide aggregation without ever accessing client data or raw rewards. At the end of each round, client $m$ sends its updated model parameters $\theta^{(m)}$ and the names and weights of its reward components $\{(r_k^{(m)}, w_k^{(m)})\}_{k}$ to the server. Using the reward \emph{names}, the server groups clients into coarse task clusters (\eg math, coding, personal assistant). Let $\mathcal{C}$ index such clusters and $\mathcal{S}_c$ denote the set of clients in cluster $c$. Within each cluster, we focus on the accuracy weight $w_0^{(m)}$, which corresponds to the primary accuracy-related objective on client $m$.

We interpret $w_0^{(m)}$ as an inverse proxy of local progress on accuracy: a smaller $w_0^{(m)}$ indicates that accuracy has been de-emphasized because it is closer to convergence. To prioritize such clients in aggregation, the server first computes inverse scores
\begin{equation}
    s_m = \frac{1}{w_0^{(m)} + \varepsilon}, \qquad m \in \mathcal{S}_c,
\end{equation}
where $\varepsilon > 0$ is a small constant, and then applies a softmax over $\{s_m\}_{m \in \mathcal{S}_c}$ to obtain normalized aggregation weights $\alpha_m$, so that clients with smaller $w_0^{(m)}$ receive larger $\alpha_m$. Within cluster $c$, model parameters and shared reward weights are aggregated as
\begin{equation}
    \theta^{(c)} = \sum_{m \in \mathcal{S}_c} \alpha_m \, \theta^{(m)}, 
    \qquad
    w_k^{(c)} = \sum_{m \in \mathcal{S}_c} \alpha_m \, w_k^{(m)}
\end{equation}
for each reward component $k$ that is common across clients in $\mathcal{S}_c$; client-specific components are not aggregated.

This yields, for each cluster $c \in \mathcal{C}$, an aggregated model $\theta^{(c)}$ and associated reward weights. In a second stage, the server aggregates across clusters in a FedAvg manner based on the total number of local training examples $N_c$:
\begin{equation}
    \theta^{\text{global}} = \sum_{c \in \mathcal{C}} \frac{N_c}{\sum_{c' \in \mathcal{C}} N_{c'}} \, \theta^{(c)}.
\end{equation}
The global model $\theta^{\text{global}}$ and the corresponding reward weights are then broadcast to clients, which resume local GRPO training in the next round.

\begin{table*}[t]
\caption{\textbf{Evaluation results on GSM8K, MATH, and HumanEval} Accuracy is reported for the global model, while values in parentheses denote the average performance of local client models. Rewards correspond to the average multi-objective reward values aggregated across objectives. Results compare FedAvg and {\n} under homogeneous and heterogeneous data–reward configurations}
\centering
\resizebox{0.99\textwidth}{!}{
\begin{tabular}{l l | c c | c c | c c}
\hline
\multirow{2}{*}{\textbf{Data+Reward}} & \multirow{2}{*}{\textbf{Method}} &
\multicolumn{2}{c|}{\textbf{GSM8K}} &
\multicolumn{2}{c|}{\textbf{MATH}} &
\multicolumn{2}{c}{\textbf{HumanEval}} \\
\cline{3-8}
 &  & \textbf{Acc} & \textbf{Rewards} &  \textbf{Acc} & \textbf{Rewards}  &  \textbf{Acc} & \textbf{Rewards} \\
\hline
\multirow{2}{*}{\textit{Homo+Homo}} & FedGRPO & 73.3(72.53) & 0.900(0.892) & 64.5(64.78) & 0.784(0.791) & \textbf{54.88}(50.37) & 0.849(0.824) \\
 & {\n} & \textbf{73.7}(72.44) & \textbf{0.902}(0.894) & \textbf{66.9}(64.55) & \textbf{0.788}(0.791) & \textbf{54.88}(50.37) & \textbf{0.850}(0.825) \\
\hline
\multirow{2}{*}{\textit{Homo+Heter}} & FedGRPO & 72.9(72.36) & 0.890(0.892) & 65.8(64.86) & 0.789(0.791) & 53.66(50.37) & 0.846(0.829) \\
 & {\n} & \textbf{74.9}(71.97) & \textbf{0.907}(0.894) & \textbf{67.0}(64.60) & \textbf{0.800}(0.800) & \textbf{55.49}(50.97) & \textbf{0.849}(0.829) \\
\hline
\multirow{2}{*}{\textit{Heter+Heter}} & FedGRPO & 72.3(69.60) & \textbf{0.897}(0.865) & 62.6(58.74) & 0.775(0.760) & 50.61(45.73) & 0.833(0.812) \\
 & {\n} & \textbf{72.4}(69.63) & \textbf{0.897}(0.872) & \textbf{64.8}(59.99) & \textbf{0.827}(0.761)  & \textbf{52.44}(49.02) & \textbf{0.839}(0.812) \\
\hline
\end{tabular}}
\label{tab:results_models_acc_rew}
\end{table*}

\section{Experiments}

\subsection{Experimental Setup}
\textbf{Models and Data.} We conduct experiments using Qwen2.5-1.5B-Instruct~\cite{qwen2025qwen25technicalreport}, an instruction-tuned LLM that is commonly used in multi-objective GRPO.
We consider three training datasets: MATH: DAPO-Math-17k-Processed (math reasoning)~\cite{yu2025dapoopensourcellmreinforcement}, GSM8K (grade-school math)~\cite{cobbe2021training}, and mbpp (python programming problems)~\cite{austin2021program}.
We evaluate on the official test sets of \textbf{MATH} and \textbf{GSM8K}, and use \textbf{HumanEval}~\cite{chen2021evaluating} for code generation.

\textbf{Federated settings.} We study three experimental categories:
\begin{itemize}
    \item \textbf{Homogeneous data + homogeneous reward (Homo+Homo).}
    All clients receive partitions of the same dataset, and share the same reward function.
    \item \textbf{Homogeneous data + heterogeneous reward (Homo+Heter).}
    All clients receive partitions of the same dataset, but use heterogeneous rewards: in addition to a shared accuracy reward, each client employs a different auxiliary reward.
    In the main experiments, we assign clients according to three pre-defined reward combinations.
    \item \textbf{Heterogeneous data + heterogeneous reward (Heter+Heter).}
    Clients are assigned different datasets (one of the three) and use dataset-specific rewards; all clients still include a shared accuracy-related reward.
\end{itemize}

For both Homo+Homo and Homo+Heter, we train on the corresponding dataset and report results on its test set.
For Heter+Heter, we train a global model over all clients and report performance separately on each benchmark test set (MATH, GSM8K, and HumanEval).

\textbf{Default hyperparameters.} All main experiments on math datasets use 10 clients for 3 global rounds. Heter+Heter use 5 clients for each training datasets, 15 in total. We select the best global checkpoint (by validation performance) as the final model for reporting. 
We use an effective batch size of 128 samples per optimization step. The learning rate is set to $2\times10^{-5}$ with a cosine decay schedule, and $\epsilon$ and $\lambda$ are set to $1\times10^{-6}$ and $0.01$, respectively. For heterogeneous data settings, we construct a hard Non-IID scenario by assigning disjoint datasets to different clients, with five clients trained on MATH, five on GSM8K, and five on MBPP, resulting in completely non-overlapping data distributions across clients. Under heterogeneous reward settings, for each dataset we consider three different reward configurations while keeping the accuracy-related reward consistent across clients; details are provided in the appendix.
To accommodate long-form reasoning traces, we set the maximum prompt length to 512 tokens and the maximum completion length to 1024 tokens. 
For GRPO sampling, 16 completions are generated per prompt, enabling the optimizer to compare multiple trajectories and compute relative advantages within each generation group. All experiments are conducted on four NVIDIA RTX 6000 GPUs. 

\textbf{Baselines.} We consider a naive federated extension of GRPO as our primary baseline, which we refer to as FedGRPO. Specifically, FedGRPO applies GRPO locally on each client and aggregates the resulting updates on the server using FedAvg. In addition, we conduct ablation studies on local adaptive objective weighting and reward-guided global aggregation to isolate their individual contributions. All hyperparameters are kept identical between FedGRPO and our method in the experiments for fair comparison. We do not directly compare against conventional FL optimization algorithms (e.g., FedProx~\cite{li2018fedprox}, FedNova~\cite{wang2020tackling}), as these methods are orthogonal to our focus on multi-objective alignment in GRPO.

\begin{figure*}[t]
    \centering  \includegraphics[width=6.2in]{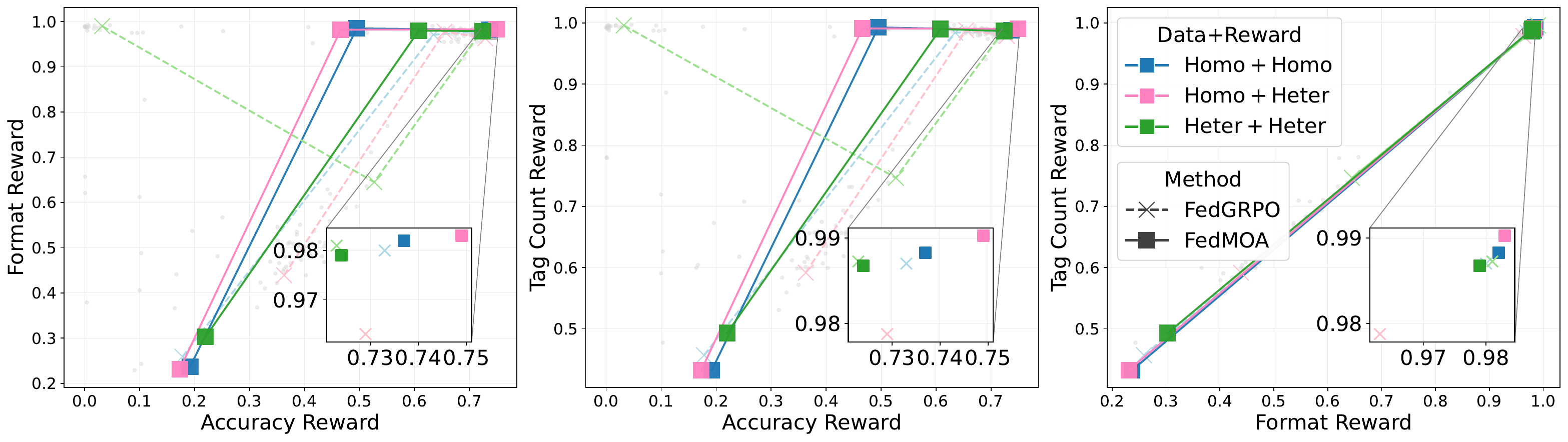}
    \centering  \includegraphics[width=6.2in]{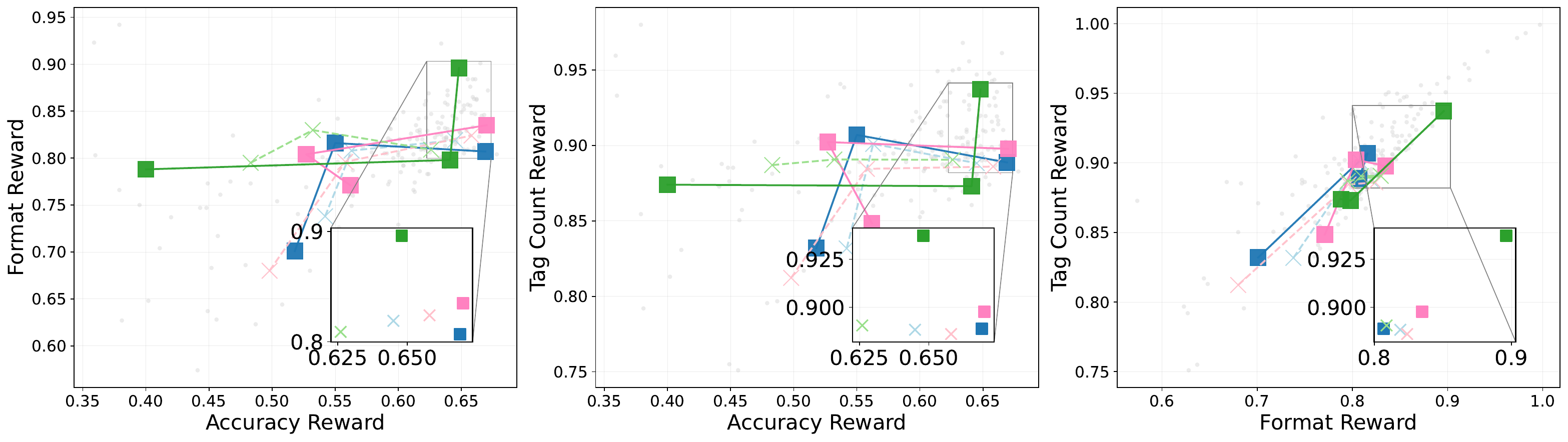}
    \caption{Trade-offs among reward components on GSM8K(upper) and MATH(lower).
Pareto frontiers are shown for format reward vs. accuracy (left), tag count reward vs. accuracy (middle), and tag count reward vs. format reward (right). Gray dots denote client-level checkpoints, while colored lines indicate server-level trajectories under different training settings. Insets zoom into the final-stage region, highlighting subtle yet consistent advantages of {\n} over FedGRPO in both homogeneous and heterogeneous settings.
}
    \label{fig:balance}
\end{figure*}

\begin{figure*}[h]
    \centering

    \begin{subfigure}{0.32\textwidth}
        \centering
        \includegraphics[width=\linewidth]{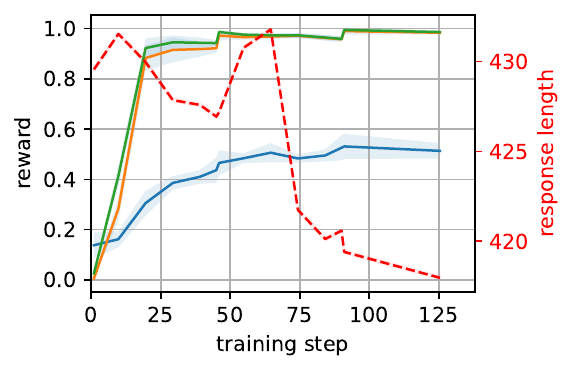}
        \caption{Homo+Homo}
    \end{subfigure}
    \hfill
    \begin{subfigure}{0.32\textwidth}
        \centering
        \includegraphics[width=\linewidth]{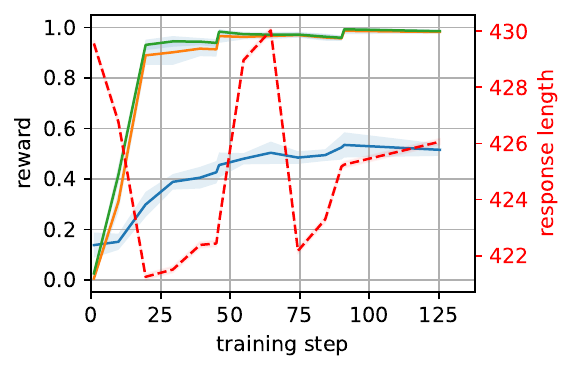}
        \caption{Homo+Heter}
    \end{subfigure}
    \hfill
    \begin{subfigure}{0.33\textwidth}
        \centering
        \includegraphics[width=\linewidth]{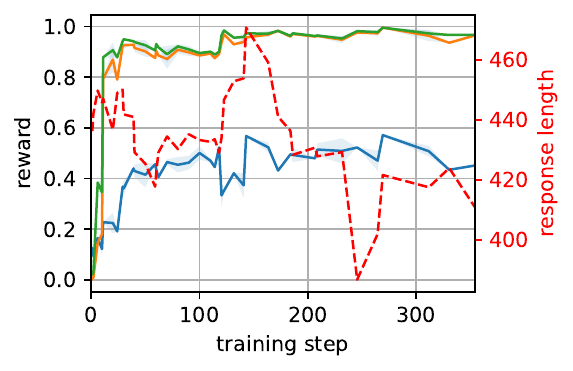}
        \caption{Heter+Heter}
    \end{subfigure}


    \begin{subfigure}{0.32\textwidth}
        \centering
        \includegraphics[width=\linewidth]{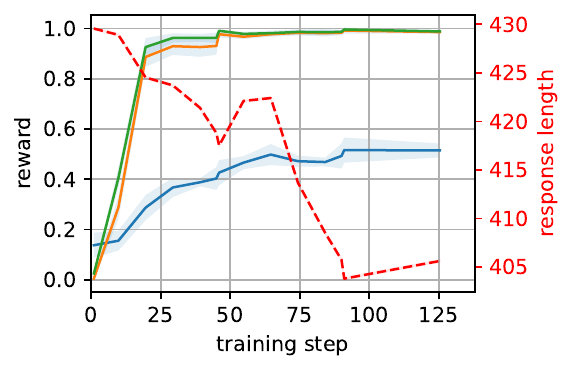}
        \caption{Homo+Homo}
    \end{subfigure}
    \hfill
    \begin{subfigure}{0.32\textwidth}
        \centering
        \includegraphics[width=\linewidth]{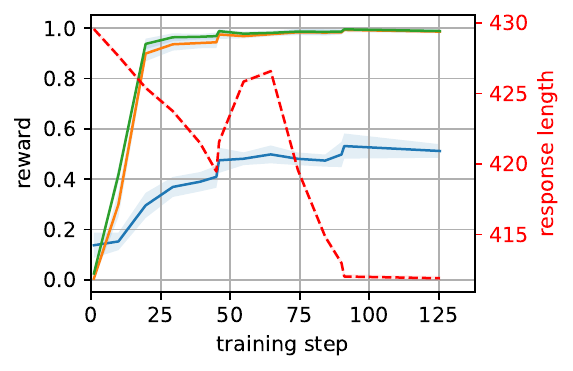}
        \caption{Homo+Heter}
    \end{subfigure}
    \hfill
    \begin{subfigure}{0.34\textwidth}
        \centering
        \includegraphics[width=\linewidth]{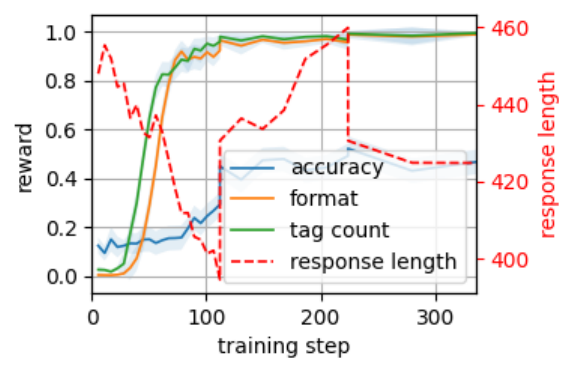}
        \caption{Heter+Heter}
    \end{subfigure}

    \caption{\textbf{Reward evolution during MATH training.}
    This figure illustrates the evolution of reward signals for {accuracy}, {format}, and {tag count}, together with the average response length (right y-axis). The top row reports results obtained with FedGRPO, while the bottom row corresponds to {\n}.}
    \label{fig:reward_dyn_math}
\end{figure*}

\subsection{Experimental Results}
To demonstrate the effectiveness of {\n}, we simulate realistic FL scenarios by varying the homogeneity of both the data distribution and the reward functions, covering Homo+Homo, Homo+Heter, and Heter+Heter. 
 We compare {\n} against FedGRPO in terms of global model accuracy and averaged multi-objective rewards.

\textbf{Results of Reasoning Performance.}
Table~\ref{tab:results_models_acc_rew} shows that {\n} improves the global model's accuracy over FedGRPO across all reported data-reward settings on GSM8K, MATH, and HumanEval. 
When all clients share identical data distribution and reward definitions, {\n} consistently outperforms FedGRPO across benchmarks. These results indicate that the proposed adaptive objective weighting and progress-aware aggregation do not degrade performance in the absence of heterogeneity. 
Under the Homo+Heter setting, the limitations of naive federated GRPO become apparent. 
{\n} achieves 74.9\% accuracy on GSM8K, outperforming FedGRPO by approximately 2\%. 
Similar global model accuracy gains are observed in HumanEval. 
These results indicate that {\n} better reconciles heterogeneous reward signals across clients.

In the most challenging Heter+Heter setting, 
{\n} demonstrates quantitative advantages over FedGRPO at both the global and client levels. 
On MATH, {\n} improves global accuracy from 62.6\% to 64.8\% and significantly increases the aggregated multi-objective reward from 0.775 to 0.827, indicating more balanced optimization under severe heterogeneity.
Notably, on HumanEval, {\n} achieves clear improvements at both the global and client levels. 
Global accuracy increases from 50.61\% to 52.44\%, while the average client-level accuracy also improves from 45.73\% to 49.02\%. These concurrent gains suggest that {\n} not only enhances the aggregated global model but also yields more effective personalized models for individual clients. 
These results together confirm that \textbf{{\n} proves particularly effective in more realistic and heterogeneous settings, consistently improving reasoning performance under both data and reward heterogeneity}.

\textbf{Results on the Multi-Objective Balance.} 
Figure~\ref{fig:balance} depicts the trade-offs among multiple reward components under different data–reward heterogeneity settings, visualized through Pareto frontiers on GSM8K (top row) and MATH (bottom row). 
Each curve traces the evolution of the global model across training, while gray points denote local model checkpoints.
Across all data–reward settings, the server curve of {\n} consistently converge to more favorable regions of the Pareto front than those of FedGRPO.
In contrast, FedGRPO often converges to regions characterized by strong auxiliary rewards but comparatively lower accuracy, reflecting suboptimal trade-offs induced by conflicting client objectives.
For example, in the GSM8K Pareto frontiers, the global model initially advances faster on easily attainable objectives such as format and tag count, while making slower progress on accuracy.
This is because improving accuracy is inherently harder, as it requires the gradual formation of coherent intermediate reasoning steps that can differ significantly across clients.
As a result, \textbf{during the early stages of training, FedGRPO can lead to unstable updates}, particularly in federated settings with heterogeneous client rewards. In contrast, adapted objective weights provide denser and smoother learning signals, enabling \textbf{{\n} to guide the global model toward a stable region of the reward space} before progressively emphasizing accuracy.

As shown in the bottom row of Figure~\ref{fig:balance}, the final server-level checkpoints produced by {\n} lie closer to the Pareto-optimal frontier in both the accuracy–format and accuracy–tag-count planes, achieving higher accuracy while preserving comparable or stronger auxiliary rewards.
Under fully homogeneous data and reward conditions, FedGRPO and {\n} exhibit largely similar trajectories, with {\n} achieving slightly better final trade-offs.
In the Heter + Heter setting, however, \textbf{FedGRPO trajectories become more scattered and make slower progress on accuracy}, whereas \textbf{{\n} maintains coordinated optimization across objectives and consistently attains higher accuracy} without sacrificing auxiliary rewards. This indicates that {\n} is more effective in strongly heterogeneous settings.

\begin{figure}[h]
    \centering
    \includegraphics[width=\columnwidth]{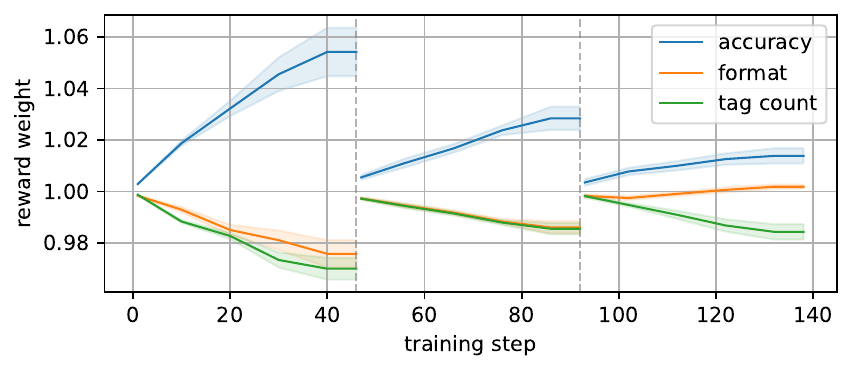}
    \includegraphics[width=\columnwidth]{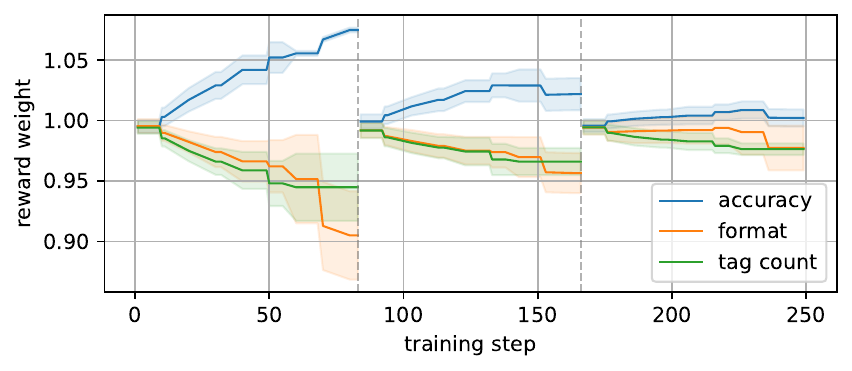}
    \caption{\textbf{Dynamic reward weight adaptation on GSM8K.}
We illustrate the evolution of reward weights for {accuracy}, {format}, and {tag count} throughout training.
The top panel shows the Homo+Homo setting, whereas the bottom panel shows the Heter+Heter setting.}
    \label{fig:reward_single_col}
\end{figure}

\textbf{Reward Dynamics and Training Stability.} 
We analyze the training dynamics of {\n} on the MATH benchmark to better understand how it stabilizes multi-objective optimization. Figure~\ref{fig:reward_dyn_math} illustrates the evolution of individual reward components together with the average response length during training. In the top row, auxiliary rewards such as {format} and {tag count} rise immediately and saturate early across all settings. By contrast, the accuracy-related reward improves more slowly and exhibits larger fluctuations, especially when reward heterogeneity is introduced. 
This pattern is most apparent in the Homo+Heter and Heter+Heter settings, where conflicting client objectives can lead to noisier and less consistent progress along the accuracy dimension.
This is most pronounced in Homo+Heter and Heter+Heter, where conflicting objectives amplify oscillations and slow accuracy improvement.
In contrast, {\n} demonstrates substantially more stable and coordinated reward dynamics. 
As auxiliary objectives gradually saturate, {\n} progressively increases emphasis on accuracy, resulting in smoother and more monotonic improvements in accuracy-related rewards, even under severe data and reward heterogeneity. 

Importantly, these improvements are not accompanied by increases in response length. 
The average response length under {\n} remains stable or decreases over the course of training, particularly in later stages. This indicates that the observed reward gains are not driven by superficial behaviors such as verbosity inflation or reward hacking, but instead reflect substantive improvements in reasoning quality and multi-objective balance.
These observations suggest that {\n} not only improves final performance but also reshapes the training dynamics of federated GRPO, enabling stable and robust multi-objective optimization under heterogeneous settings.

\textbf{Adaptive Objective Weighting Dynamics.}
As shown in Figure~\ref{fig:reward_single_col}, reward weights on GSM8K exhibit distinct evolution patterns under different data–reward heterogeneity settings. 
In the Homo+Homo setting, reward weights evolve smoothly after being reset at each round boundary. 
As training progresses, the weight assigned to accuracy gradually increases, while auxiliary objectives such as format and tag count decrease or stabilize once they saturate early. 
Notably, the rate of increase in the accuracy weight diminishes over successive rounds, suggesting a gradual convergence toward a balanced objective configuration. 
This pattern reflects a gradual shift in emphasis from easier auxiliary objectives toward the primary reasoning objective as auxiliary performance becomes sufficient.

In the Heter+Heter setting, where both data distributions and reward functions are heterogeneous, the reward weights exhibit substantially higher variability; in particular, adaptive reweighting becomes more actively engaged to maintain a balance between accuracy and auxiliary rewards.
As a result, {\n} maintains stable multi-objective optimization dynamics even in Non-IID settings, where reward availability and client participation are inherently variable.

\textbf{Ablation Studies.}
Table~\ref{tab:ablation_math} presents an ablation study on MATH under the Heter+Heter setting, analyzing the effects of adaptive objective weighting ($\lambda$) and accuracy-aware aggregation ($\alpha$).
When both components are disabled, the method reduces to FedGRPO, yielding a global accuracy of 62.6\% and an aggregated reward of 0.775.
Enabling accuracy-aware aggregation alone ($\alpha$ ON, $\lambda$ OFF) improves global accuracy to 64.0\% but leaves aggregated reward of essentially unchanged (0.776), suggesting that aggregation by itself provides only limited benefit under heterogeneous reward signals. 
In contrast, combining adaptive objective weighting with accuracy-award aggregation ($\alpha$ ON, $\lambda$ ON) further improves global accuracy to 64.8\%  and substantially increases the aggregated reward to 0.827, indicating more balanced multi-objective optimization. 
These results suggest that client-side reweighting helps mitigate local objective interference, allowing accuracy-aware aggregation to more effectively prioritize high-quality updates under heterogeneous data-reward settings. 

\section{Related Work}
\paragraph{LLM Federated Learning.}
Federated learning (FL) was popularized by FedAvg~\cite{mcmahan2017fedavg}, followed by extensive studies addressing client heterogeneity and optimization stability, including FedProx, SCAFFOLD, FedOpt, and FedExp~\cite{li2018fedprox,karimireddy2020scaffold,asad2020fedopt,jhunjhunwala2023fedexp}. When extending FL to large language models (LLMs), recent work has primarily focused on \emph{communication- and memory-efficient} federated adaptation, most notably via parameter-efficient fine-tuning (PEFT) techniques such as LoRA or adapters together with FL-specific aggregation schemes~\cite{zhang2024towards,sun2024improving,wang2024flora}. To further strengthen privacy guarantees in federated LLM fine-tuning, several approaches restrict the training interface, including prompt-based methods~\cite{sun2023fedbpt} and black-box fine-tuning in edge settings~\cite{wang2025prada}.  
Federated reinforcement learning has also been studied~\cite{qi2021federated,pinto2023federated,rengarajan2024federated}, but existing works are largely limited to small or linear models~\cite{xiong2024linear} and conventional RL paradigms, such as Q-learning~\cite{zheng2023federated} or RLHF-style optimization~\cite{fan2024fedrlhf}. In contrast, the integration of FL with multi-objective GRPO for large-scale language models remains unexplored.

\paragraph{LLM Reinforcement Learning with GRPO.}
Reinforcement learning for LLM alignment is commonly built on on-policy policy-gradient methods, most notably PPO~\cite{schulman2017proximal}, with substantial effort devoted to reducing training cost and improving stability under preference- or verifier-based supervision. GRPO was proposed as a PPO-style alternative that eliminates the need for an explicit critic by computing \emph{relative} advantages within a group of sampled responses conditioned on the same prompt, and was successfully applied in DeepSeekMath to scale reasoning-oriented RL under tight memory budgets~\cite{guo2025deepseek}. Subsequent work has further examined GRPO’s objective formulation and estimation properties, and proposed extensions such as off-policy variants and guided or adaptive group comparisons to improve sampling efficiency and robustness~\cite{mroueh2025revisiting,guo2025g}.  
This paper offers a new perspective on GRPO by studying it in a distributed and privacy-preserving FL setting, where optimization and alignment must be coordinated across heterogeneous clients with limited information sharing.



\begin{table}[t]
\centering
\caption{Ablation study on MATH(Heter+Heter setting). We analyze the effects of adaptive objective
weighting ($\lambda$) and accuracy-aware aggregation ($\alpha$). The top row is FedGRPO.}
\label{tab:ablation_math}
\begin{tabular}{c l | c c}
\toprule

\textbf{ $\lambda$} 
& \textbf{ $\alpha$}  & \textbf{Acc} & \textbf{Reward} \\
\hline
OFF & OFF 
& 62.6(58.74) & 0.775(0.760)  \\
\hline
OFF & ON 
 & 64.0(58.91) & 0.776(0.760) \\
\hline
ON & ON  & \textbf{64.8}(59.99)  & \textbf{0.827}(0.761)  \\
\hline
\end{tabular}
\end{table}

\section{Conclusion}
This work investigated federated GRPO under heterogeneous multi-objective rewards and identified key challenges related to optimization imbalance, personalization, and training efficiency. To address these challenges, we proposed {\n}, which combines client-side adaptive objective weighting with task-aware and accuracy-aware aggregation at the server level.
Experiments on mathematical reasoning and code generation benchmarks demonstrate that {\n} consistently outperforms standard federated baselines, improving global model performance, enhancing per-client personalization, and achieving more balanced multi-objective trade-offs under both homogeneous and heterogeneous settings. In addition, {\n} leads to more stable training dynamics across clients with diverse objectives, particularly in highly Non-IID regimes. Overall, these results highlight federated GRPO as a practical approach for privacy-preserving personalization of large language models and motivate future work on scaling to larger models and more complex objective structures.


\newpage

\section*{Impact Statements}
This paper presents work whose goal is to advance the field of machine learning. In particular, it studies federated reinforcement learning methods for aligning large language models under heterogeneous objectives. While such techniques may enable more effective and personalized model training in privacy-preserving settings, we do not foresee any immediate negative ethical or societal consequences specific to this work beyond those generally associated with large language models and federated learning systems.

\bibliography{example_paper}
\bibliographystyle{icml2026}

\newpage
\appendix
\onecolumn
\section{Additional Experimental Settings}
\subsection{Heterogeneous reward configurations}
To model \emph{reward heterogeneity}, we assign three distinct multi-objective reward configurations to clients within each dataset, while keeping the \texttt{accuracy} reward identical across clients. This isolates the effect of heterogeneous auxiliary rewards from differences in task correctness. Table~\ref{tab:heter_reward_configs} summarizes the configurations; exact formulations and coefficients are provided in the appendix.

\begin{table}[h]
\caption{Heterogeneous reward configurations used in our experiments. The \texttt{accuracy} reward is shared across clients, while auxiliary rewards vary across configurations to induce reward heterogeneity. See the appendix for exact formulations and coefficients.}
\centering
\small
\setlength{\tabcolsep}{6pt}
\begin{tabular}{lccc}
\toprule
\textbf{Dataset} & \textbf{Config A} & \textbf{Config B} & \textbf{Config C} \\
\midrule
\multirow{3}{*}{MATH}
& \texttt{accuracy} & \texttt{accuracy} & \texttt{accuracy} \\
& \texttt{format} & \texttt{format} & \texttt{tag count} \\
& \texttt{tag count} & \texttt{-}  & \texttt{-} \\
\midrule
\multirow{3}{*}{GSM8K}
& \texttt{accuracy} & \texttt{accuracy} & \texttt{accuracy} \\
& \texttt{format} & \texttt{format} & \texttt{tag count} \\
& \texttt{tag count} & \texttt{-}  & \texttt{-} \\
\midrule
\multirow{3}{*}{MBPP}
& \texttt{accuracy} & \texttt{accuracy} & \texttt{accuracy} \\
& \texttt{code\_format} & \texttt{code\_format} & \texttt{code\_format} \\
& \texttt{tag count} & \texttt{length} & \texttt{repetition\_penalty} \\
\bottomrule
\end{tabular}
\label{tab:heter_reward_configs}
\end{table}

\end{document}